\documentclass[10pt,twocolumn,letterpaper]{article}

\usepackage{wacv}
\usepackage{subfigure}
\usepackage{graphicx}
\usepackage{amsmath}
\usepackage{amssymb}
\usepackage{enumitem}
\usepackage{algpseudocode}
\usepackage{color}
\usepackage{amsmath}
\usepackage{amssymb}
\wacvfinalcopy 

\ifwacvfinal\fi
\begin{document}

\title{Region Graph Based Method for Multi-Object Detection and Tracking using Depth Cameras\thanks{Appearing in Proc. IEEE Winter Conference on Applications of Computer Vision, USA, 2016}}

\author{Sachin Mehta\thanks{This work was done by author before joining University of Washington.} \\
University of Washington\\
{\tt\small sacmehta@uw.edu}
\and
Balakrishnan Prabhakaran \\
University of Texas at Dallas\\
{\tt\small bprabhakaran@utdallas.edu}
}

\maketitle
\begin{abstract}
  In this paper, we propose a multi-object detection and tracking method using depth cameras. Depth maps are very noisy and obscure in object detection. We first propose a region-based method to suppress high magnitude noise which cannot be filtered using spatial filters. Second, the proposed method detect Region of Interests by temporal learning which are then tracked using weighted graph-based approach.  We demonstrate the performance of the proposed method on standard depth camera datasets with and without object occlusions. Experimental results show that the proposed method is able to suppress high magnitude noise in depth maps and detect/track the objects (with and without occlusion).
\end{abstract}


\section{Introduction}
\label{sec:introduction}
The recent emergence of cost-effective depth sensors have triggered significant attention of researchers within computer vision and robotics community. Depth sensors have their own advantages over visible light cameras. Depth images are insensitive to variation in texture and illumination. Moreover, depth images represent 3D structural information reflecting shape cues and geometry. 

There is a growing body of research on topics such as human detection and human tracking using 3D information. Ikemura et al. \cite{Ikemura2011:eke} proposed a method for human detection. The method extracts relational depth similarity features and constructs a classifier which is then used for detecting humans. Xia et al. \cite{Xia2011:eke} proposed human detection method in indoor environment using depth information. The method first identifies the possible regions that may contain humans which are then verified using 3D head model. Zhang et al. \cite{Zhang2012:eke} explored the Kinect for detecting falls in elderly people. The method eventually adopt a known background subtraction method for detecting people and analyze the trajectory to detect the fall. 

Instead of using only depth maps, some researchers exploited both RGB and depth channels for human detection and tracking. Han et al. \cite{Han2012:eke} proposed a method for detecting and tracking humans using both RGB and depth channels. The method locates the object in depth map and then extracts the corresponding visual features from the RGB data. These visual features are then used for tracking objects in successive frames of RGB data. Spinello et al. \cite{Spinello2011:eke} extends the idea of  Histogram of Oriented Gradients (HOG) and proposed a conceptually similar Histogram of Oriented Depths (HOD) method for human detection. Enzweiler et al. \cite{Enzweiler2010:eke} developed a stereo system which combines intensity images, stereo disparity maps, and optical flow for detecting and tracking people. Song et al. \cite{Shuran2013:eke} developed a RGB-D dataset for tracking objects. They implemented and tested different algorithms, such as HOG with Optical flow, on RGB-D data for tracking objects.

Graph-based object tracking methods have been proposed in past for RGB videos. Gomilla and Meyer \cite{Gomila2003:eke} proposed a graph-based object tracking method whereby each frame is represented as a region adjacency graph and thus, tracking becomes a graph-matching problem. Wang and Nevatia \cite{Wang2013:eke} proposed a tracking method based on superpixels. Constellation appearance model was constructed based on visual features extracted from superpixels and then tracking is performed using Dynamic Bayesian Network. Recently, Yang et al. \cite{Yang2014:eke} proposed a tracking method using superpixels. They used mid level cues for distinguishing between target and the background. Tracking is then achieved by formulating a map between target and background. 

In this paper, we have successfully applied region graph based approach for object detection and tracking in depth videos. The proposed approach shows a significant improvement of about 9\% over the existing methods. Further, depth maps are very noisy and contain high magnitude noise which is difficult to remove using spatial filter. We proposed a method to remove such high magnitude noise from depth maps and have seen a significant improvement of about 23\% in tracking results over the tracking results obtained without noise suppression.

Rest of the paper is organized as: The proposed method for noise suppression, ROI detection, and tracking is discussed in Section \ref{sec:proposedMethod}. Experimental results are presented in Section \ref{sec:expRes} while conclusions are drawn in Section \ref{sec:conclusion}.

\section{Proposed Method}
\label{sec:proposedMethod}
In this section, region graph based approach for ROI detection and tracking is discussed in detail.
\subsection{Noise Suppression}
\begin{figure}[t!]
\centering
\subfigure[Original RGB Image]{\label{fig:orig}\includegraphics[width=4cm]{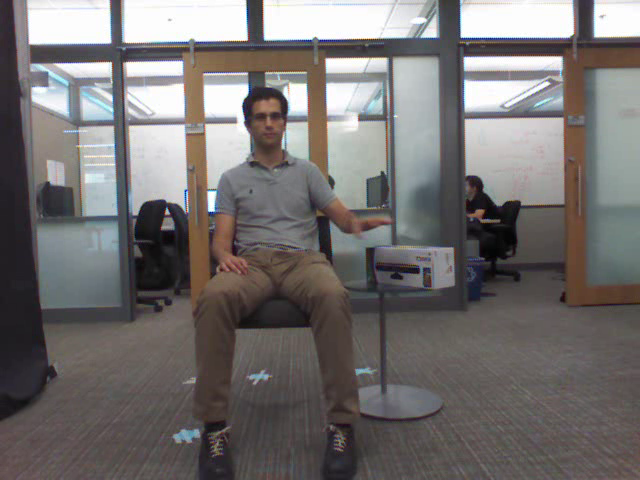}}%
\vspace{0.05cm}
\subfigure[Original Depth Map]{\label{fig:orig}\includegraphics[width=4cm]{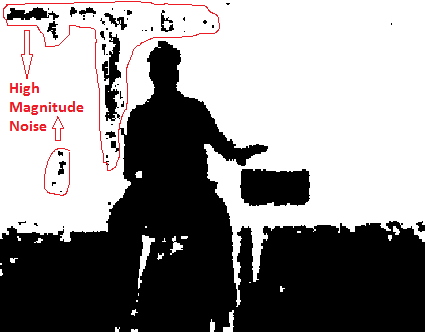}}%
\hspace{0.05cm}
\subfigure[Xia's \cite{Xia2013:eke} Method]{\label{fig:cvpr}\includegraphics[width=4cm]{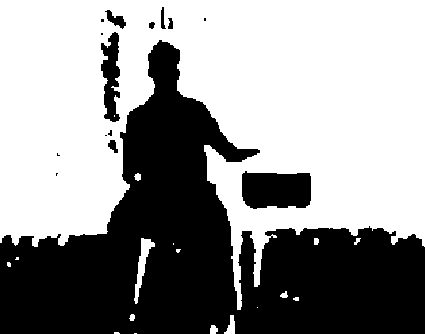}}%
\vspace{0.05cm}
\subfigure[The Proposed Method]{\label{fig:prop}\includegraphics[width=4cm]{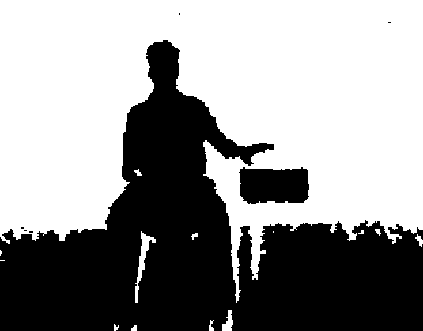}}%
\caption{Noise Filtering Example. For the sake of understanding, we have shown the depth maps as binary images.}
\label{fig:noiseCompare}
\end{figure}

Noise in depth videos can be classified into three categories: (i) noise from sensors $N_S$, (ii) noise from boundary of objects $N_B$, and (iii) holes in depth map $N_H$ (caused by fast movements, random surfaces, etc.) \cite{Xia2013:eke}. Noise occurred due to variation in sensors is evenly distributed throughout the depth map and is of low magnitude. Such  type of noise can be easily removed using spatial filters such as Gaussian filter. On the other hand, noise occurring due to boundary of objects and holes in depth map is not evenly distributed and their magnitude is quite high in contrast to noise arising from sensors. Such type of noise is difficult to remove using spatial filters. Based on temporal learning, Xia et al. \cite{Xia2013:eke} proposed a filtering algorithm for removing such high magnitude noise. Figure \ref{fig:cvpr} shows a snapshot of the filtered depth map using Xia et al.'s \cite{Xia2013:eke} method. The method is able to remove some part of high magnitude noise but not all. In this section, a method is proposed to remove high magnitude noise arising from boundary of objects and holes in depth map.

\begin{figure}[t!]
\centering
\subfigure[]{\includegraphics[width=0.38\columnwidth]{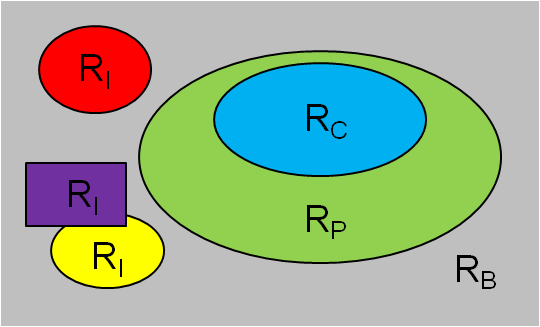}}%
\hspace{0.5cm}
\subfigure[]{\includegraphics[width=0.38\columnwidth]{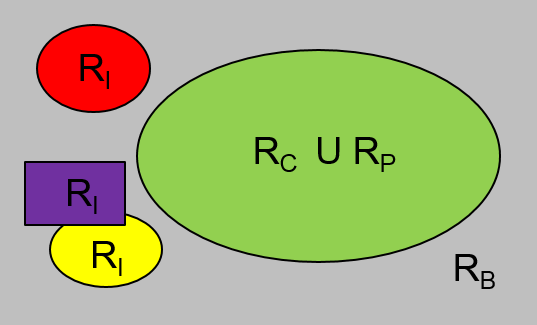}}%
\caption{Region Categorization: (a) Without merging $R_P$ and $R_C$ region (b) With merging $R_P$ and $R_C$ region}
\label{fig:regionCategory}
\end{figure}

The proposed method first applies a 2D Gaussian smoothing filter to remove noise arriving from sensors. Then, the proposed method segments the depth map into regions using morphological watershed segmentation (MWS) proposed by Meyer et al. \cite{Meyer1990:eke}\footnote{Any other image segmentation algorithm can be used.}. Let us assume that $R$ closed regions and $R_B$ background regions are detected in spatially filtered depth map $D_S$ using MWS\footnote{We assume that depth map has at least one region.}. Background regions $R_B$ are the ones whose at least $P$ boundary points lie on the border of depth map. Some of the closed regions might enclose other closed regions. We represent enclosed regions as $R_C \subset R$, enclosing regions as $R_P \subset R$, and the remaining closed regions as independent regions $R_I \subset R$ such that $ R = R_C \cup R_P \cup R_I$. Figure \ref{fig:regionCategory} shows different possible regions. Region $R_C$ and $R_P$ can be merged and the merging decision is solely dependent upon application. For instance, region $R_C$ and $R_P$ can not be merged for human part-based tracking while for complete human tracking, $R_C$ and $R_P$ can be merged. In the proposed method, we are interested in complete object tracking in contrast to part-based tracking and hence, we merge regions $R_C$ and $R_P$.

\begin{figure}[t!]
\centering
\includegraphics[width=0.7\columnwidth]{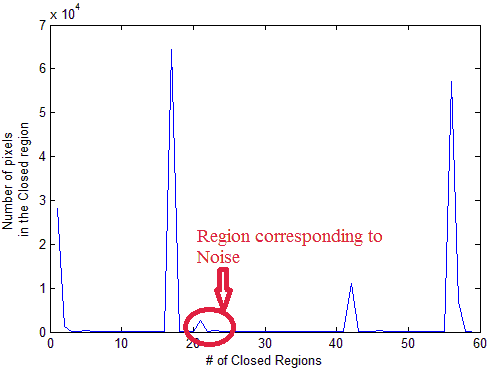}
\caption{Area of closed regions contained in Figure \ref{fig:orig}}
\label{fig:noiseMagnitude}
\end{figure}

Now, the proposed method computes the area $A_R$ of each closed region $R$ as: $A_R[i] = \sum \sum R[i](x,y)$.
Noise arising from boundary of objects $N_B$ and holes $N_H$ in depth map $D$ will also create regions. Though magnitude of $N_B$ and $N_H$ will be higher in comparison to $N_S$, it has lower magnitude in contrast to actual object size. Figure \ref{fig:noiseMagnitude} shows the plot of area of closed regions contained in Figure \ref{fig:orig}. In Figure \ref{fig:noiseMagnitude}, we can see that the area of regions corresponding to noise is quite less (No. of pixels lies between $0.01$ $\times 10^4$ and $0.3$ $\times 10^4$) in comparison to regions corresponding to object size (No. of pixels lies between $1$ $\times 10^4$ and $7$ $\times 10^4$).  Threshold based approach is then applied to remove noise from spatially filtered depth map $D_S$ as:
\begin{equation}
 R[i]= \left\{
\begin{array}{l}
R[i], \quad A_R[i] > \tau \\
0, \quad otherwise
\end{array}
\right.
\end{equation}
where $\tau$ is the threshold for eliminating the noise arising from boundary of objects and holes in depth map.  $\tau$ is a function of region area and can be computed as $\tau = \frac{|| A_R[i]||_1}{n}$, $1\leq i\leq n$. 

Figure \ref{fig:prop} shows an example of noise suppression in depth map using the proposed method. From Figure \ref{fig:prop}, it is clear that proposed method is able to remove the noise ($N_B$, $N_S$, and $N_H$) effectively. 
\subsection{Good features to track}
\label{ssec:featureToTrack}
 Object traverses a very limited space from time $t$ to time $(t+1)$. We use this important clue to map regions between two consecutive frames and identify ROI. Let $G_t$ and $G_{t+1}$ be the undirected graphs for frame at time $t$ and $t+1$. Displacement between two arbitrary nodes, $i_0\ \in\ G_t$ and $j_0\ \in\ G_{t+1}$, with region $R[i_0]$ and $R[j_0]$ respectively is defined as: $\mathcal{D}(j_o \rightarrow i_0) = R[j_0] \setminus R[i_0]$. This measure tells us how much a region has traversed from time $t$ to $(t+1)$. Furthermore, region mapping process is speeded-up by including the threshold $\delta$. If $\mathcal{D}(j_o \rightarrow i_0) > \delta$, then we map the regions $R[j_0]$ and $R[i_0]$\footnote{Empirical value of $\delta$ in our experiments is 80.}.

Besides region mapping, the proposed method simultaneously do temporal tracking of ``mapped regions'' in first $K$ depth maps to determine ROI. ROI can be either growing region (region whose area changes, as in case of hand waving) or shrinking region (region whose area changes, as in case of zooming out) or moving region (region whose position changes, as in case of rolling ball) or combination of growing, shrinking, and movement (as in case of box lifting). The proposed method computes $||\mathcal{D}(j_o \rightarrow i_0)||_1$ from mapped regions of first $K$ frames. If $||\mathcal{D}(j_o \rightarrow i_0)||_1$ is greater than a particular threshold value\footnote{Empirical value is $70$}, then we mark the region $R[j_o]$ as ROI.

\begin{figure}[h!]
\centering
\includegraphics[width=0.15\columnwidth]{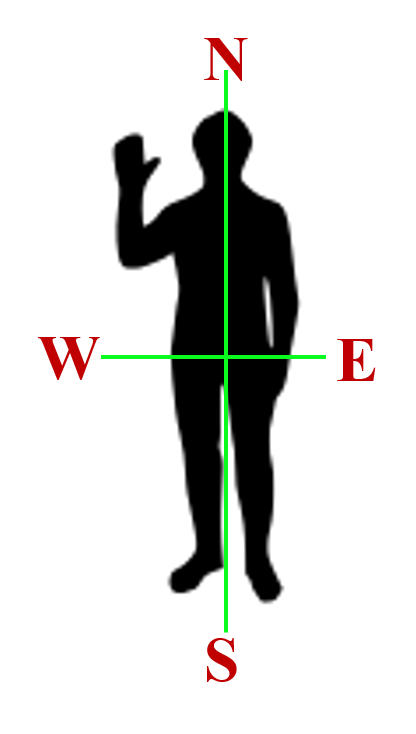}
\caption{Four Point Cardinal System}
\label{fig:boundaryPointDetection}
\end{figure}
\begin{figure}[h!]
\centering
\subfigure[]{\label{fig:frameA}\includegraphics[width=0.2\columnwidth]{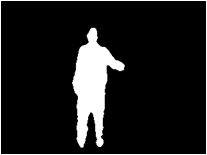}}%
\hspace{0.1\columnwidth}
\subfigure[]{\label{fig:frameB}\includegraphics[width=0.2\columnwidth]{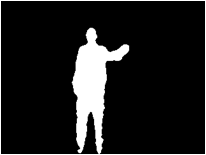}}%
\hspace{0.1\columnwidth}
\subfigure[]{\label{fig:diffAB}\includegraphics[width=0.2\columnwidth]{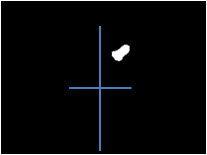}}%
\caption{Determining direction (a) Depth frame at time $t$ (direction of hand is east), (b) depth frame at time $t+1$ (direction of hand is North-East) and (c) Difference between (a) and (b)}
\label{fig:directionDeter}
\end{figure}

As the proposed method is region-based, direction of object movement is an important clue and can help in optimizing the performance of the proposed method while searching the ROI in next depth frame. We use a cardinal system to determine the direction of ROI movement. Cardinal system can be 4-point or 8-point or 16-point. In our experiments, we have used 4-point cardinal system, shown in Figure \ref{fig:boundaryPointDetection}. For determining the direction of ROI, the proposed method computes difference between the mapped regions of two consecutive depth maps $\mathcal{D}(j_o \rightarrow i_0)$ and then decides the direction of region movement using the four point cardinal system. It is to be noted that the direction is computed with respect to the center of the region of depth frame at time $(t+1)$ i.e. $R[j_0]$. Figure \ref{fig:frameA} and Figure \ref{fig:frameB} shows snapshots of two consecutive depth maps where person is waving a hand while the difference between these depth maps is shown in Figure \ref{fig:diffAB}. From Figure \ref{fig:diffAB}, we can see that the direction of hand movement is North-East. Given this information, we can predict that the hand movement in the next depth frame will be either in East or North direction and we do not need to search in South and West direction.

\subsection{Tracking}
\label{ssec:tracking}
\begin{figure}[t!]
\centering
\subfigure[]{\label{fig:unDir}\includegraphics[width=0.4\columnwidth]{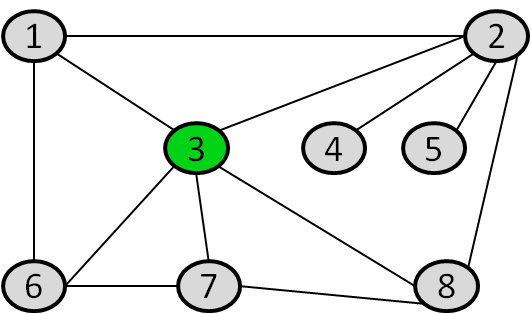}}%
\hspace{0.5cm}
\subfigure[]{\label{fig:unDirDist}\includegraphics[width=0.4\columnwidth]{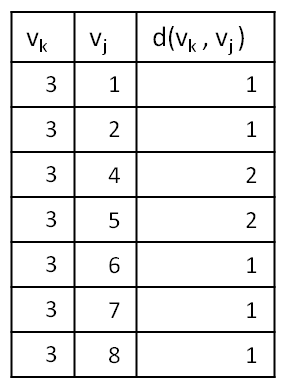}}%
\vspace{0.05cm}
\subfigure[]{\label{fig:unDirWt}\includegraphics[width=0.45\columnwidth]{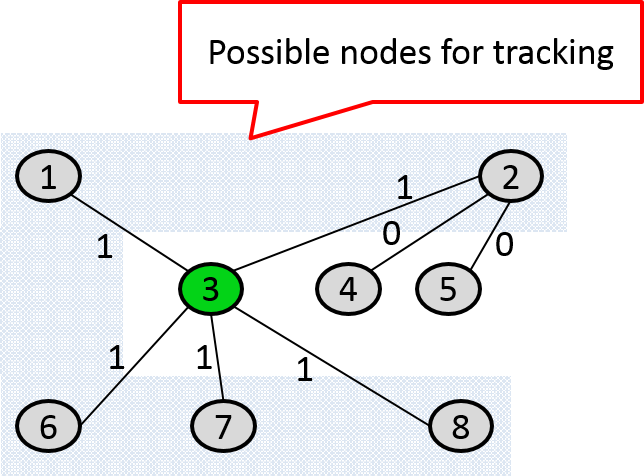}}%
\hspace{0.5cm}
\subfigure[]{\label{fig:possibleNodes}\includegraphics[width=0.45\columnwidth]{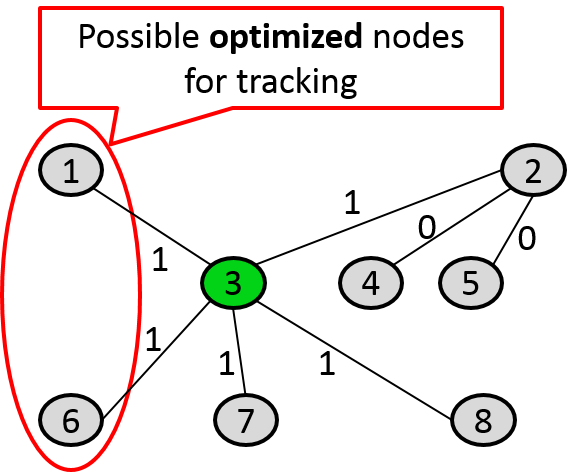}}%
\caption{(a) Graph $G$, (b) Node table for Graph $G$ shown in (a), (c) Weighted graph, and (d) Optimized weighted graph for object tracking}
\label{fig:undirGraph}
\end{figure}

The proposed method starts tracking ROI from $(K+1)^{th}$ depth map ($K$ depth maps are required for ROI determination). In order to track ROI, the proposed method creates a weighted graph for each ROI. Let $V$ and $E$ denotes the number of nodes and edges in graph $G$ of $(K+1)^{th}$ frame, where each node $v_i\ \in\ V$ corresponds to each region in the depth map (set of elements to be tracked) and edges $(v_i,\ v_j)\ \in\ E$ corresponding to neighbouring regions\footnote{We draw the boundary between objects using distance transform.}. Let us assume that node $v_k$ represents ROI. Now, a node table is constructed for node $v_k$. Node table contains the shortest distance $d$ between ROI and the remaining regions. Now, the weights are assigned to edge $(v_k,\ v_j)\ \in\ E$ as:
\begin{equation}
w(v_k,\ v_j) = \left\{
\begin{array}{l}
1, \quad d(v_k,\ v_j) = 1 \\
0, \quad otherwise
\end{array}
\right.
\label{eq:weightAssign}
\end{equation}
Figure \ref{fig:undirGraph} shows an example of constructing a node table and assigning weights $w$ to the edges of a graph $G$. To illustrate, let us assume that $v_3$ is the ROI (represented in green color). Now, the proposed method computes the distance between node $v_3$ and remaining nodes of graph $G$ and constructs a node table, shown in Figure \ref{fig:unDirDist}. Based on the values of node table and Eq. \ref{eq:weightAssign}, weights are assigned to the edges (see Figure \ref{fig:unDirWt}). When an object moves, set of attributes (such as area and perimeter) of adjoining regions are impacted. In other words, when any movement occurs in region corresponding to node $v_3$, attributes of the edges $(v_3, v_j)\ \in\ E$ with weight $w(v_3, v_j)=1$ will be affected most. Movement in node $v_3$ will have an impact on the attributes of nodes $v_1$, $v_2$, $v_6$, $v_7$, and $v_8$ most. Since the attributes of nodes $v_4$ and $v_5$ will have the least impact due to any movement in $v_3$, we do not have to consider these nodes for tracking. This ultimately helps in reducing the searching area for object tracking. In order to further minimize the searching area, the proposed method utilizes predicted direction of the ROI. If the ROI is moving towards West, for instance, then regions with edges $(v_3, v_j)\ \in\ E$ and weight $w(v_3, v_j)\ =\ 1$ in the West direction are most probable regions for tracking. In our example, nodes $v_1$ and $v_6$ are in West of $v_3$ and hence, we need to search in regions corresponding to these nodes. 

Objects can appear or disappear in a scene. The proposed method runs a background thread at a constant frame interval to check if any ROI is appearing or disappearing in the scene. This background thread executes ROI detection method discussed in Section \ref{ssec:featureToTrack} and if there is any change in number of ROIs, it updates the object tracking thread.  In case number of ROIs are more than one, the proposed method initiates different foreground thread to track each ROI. Value of constant frame interval in our experiments is  equal to number of frames required to detect ROI i.e. $K$. 

\subsection{Multi-object Tracking and Occlusion Detection}
Multiple objects or ROIs in a scene can occlude if they are moving towards each other. Occlusion can be detected based on ROI's size and the distance between ROI's. Let $A$ and $B$ be two ROI's detected and $d_e$ be the euclidean distance between $A$ and $B$. If $d_{e}=0$, then the two ROI's are about to occlude. If the area of $A$ is changing drastically while the change in area of $B$ is either zero or slight, then we mark ROI $A$ as \textit{occludee}.
  
Let us say that there are $N$ ROI's with level sets $\theta_i$. The Euclidean distance $d_e$ between ROI $O_i$ and ROI $O_j$ can be obtained as: $d_{e} = arg\ min\ \theta_i(\theta_j), i \ne j$. Now, we compute change in the area of ROI $O$ as: $\Delta A_R=| A_R^t \setminus A_R^{t-1} |$, where $A_R^t$ denotes the area of ROI $O$ at time $t$.

The proposed method then computes the occlusion detection parameter $OD$ between ROI $O_i$ and ROI $O_j$ as: $OD_{i,j} = \frac{\Delta A_R}{exp^{-|d_{e}|} + 1}$. If $OD_{i,j} < \iota$, then we say that occlusion occurs where $\iota$ is the threshold parameter for occlusion detection. We conducted experiments on samples with two or more humans to adjust the value of $\iota$. We varied the value of $\iota$ from 0 to 1 in our experiments. When we set $\iota<0.5$, we were able to detect the occlusion as soon as it occurs. However, when we set $\iota \ge 0.5$, we were able to detect the occlusion but only when at least 25\% of the region of two humans were overlapping. Hence, we set $\iota=0.4$ in our experiments. 

\begin{figure}[t!]
\centering
\subfigure[]{\includegraphics[height=2cm]{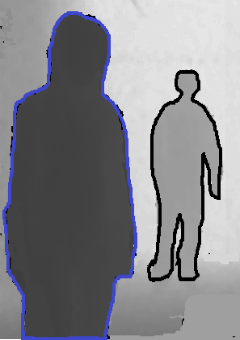}}%
\hspace{0.5cm}
\subfigure[]{\includegraphics[height=2cm]{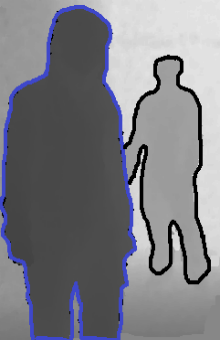}}%
\hspace{0.5cm}
\subfigure[]{\includegraphics[height=2cm]{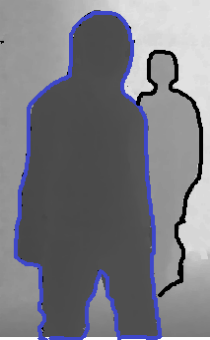}}%
\caption{A sequence in which two humans occlude each other ($\iota = 0.4$)}
\label{fig:occHandle}
\end{figure}

Figure \ref{fig:occHandle} shows an occlusion example where two humans occlude each other (in this sequence, both the humans are in motion). Before the occlusion, the boundaries of the humans are complete. During occlusion, two humans start overlapping each other (Figure \ref{fig:occHandle}(b)) and hence, the boundaries are broken. At this point, the distance between two humans is very less ($d_{e} \approx 0$). 

\section{Experimental Results}
\label{sec:expRes}
We evaluated the performance of the proposed method on $1000^+$ videos taken from standard datasets (MSR \cite{Li2010:eke} \cite{Jiang2012b:eke}, UT Austin \cite{Xia2011:eke}, Princeton University \cite{Shuran2013:eke}, IROS \cite{Spinello2011:eke}, and RWTH \cite{Rafi2015:eke}). Videos in these datasets differ in terms of pose, background, actions, etc. We conducted experiments on a computer having Intel i5-2410M 2.30 GHz processor and 6GB DDR3 RAM.

\begin{figure}[b!]
\centering
\includegraphics[width=0.8\columnwidth]{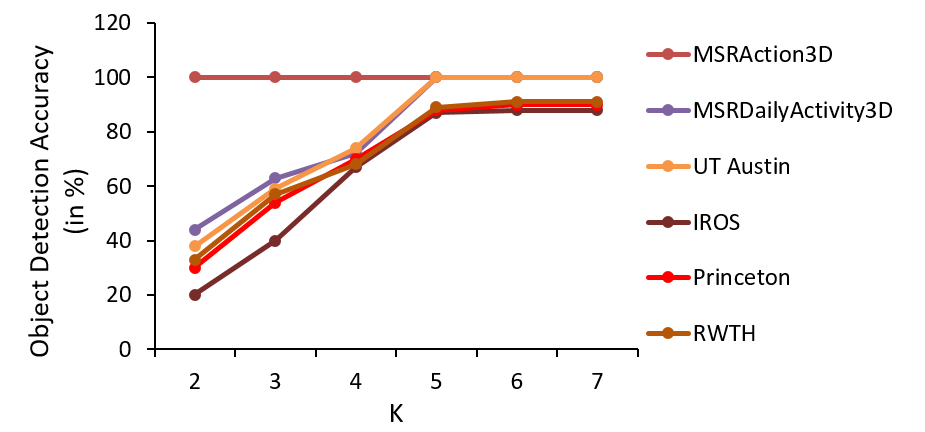}
\caption{Impact of $K$ on object detection accuracy}
\label{fig:objectAccuracy}
\end{figure}

Since the proposed method uses $K$ depth maps for detecting ROI,  it should be considered as an independent variable to conduct a fair study. We conducted experiments to adjust the value of $K$ and results are shown in Figure \ref{fig:objectAccuracy}. From Figure \ref{fig:objectAccuracy}, we can see that the performance plateaus after $K=5$. Hence, we used $K=5$ in our experiments.

\subsection{Metrics for evaluation}
\label{ssec:metrics}
We quantify the performance of the proposed method using two metrics:
\begin{itemize}
\item \textbf{$F_1$ Score for object detection}: We used this metric for determining the accuracy of the proposed object detection method \cite{Kootstra2011:eke}. $F_1$ score combines both precision and recall and is defined as: $F_1\ score = \frac{2 \times \mathcal{P} \times \mathcal{R}}{ \mathcal{P} + \mathcal{R}}$, where $\mathcal{P} = \frac{TP}{TP+FP}$ and $\mathcal{R} = \frac{TP}{TP +FN}$. Here, $\mathcal{P}$ represents precision, $\mathcal{R}$ represents recall, $TN$ represents true negative samples, $TP$ represents true positive samples, and $FN$ represents false negative samples.
\item \textbf{Success rate (SR) for tracking}: For quantitative evaluation of our tracking algorithm, we employed the criterion used in PASCAL VOC challenge \cite{Everingham2010:eke}. If $r > 0.5$, then frame is tracked successfully. Here, $r = \frac{\mathcal{T} \cap \mathcal{G}}{\mathcal{T} \cup \mathcal{G}}$ denotes the overlap ratio, $\mathcal{T}$ denotes the tracked region, and $\mathcal{G}$ denotes the ground truth.
\end{itemize}
To compare the results with previous methods, we either take the reported best results or carefully select the parameters with the provided source code.

\subsection{Object Detection Results}
\label{ssec:detectResult}
\begin{figure}[b!]
\small
\centering
\subfigure[]{\includegraphics[width=0.2\columnwidth]{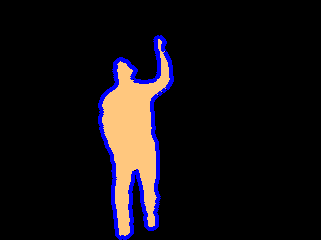}}%
\hspace{0.01cm}
\subfigure[]{\label{fig:boxTh}\includegraphics[width=0.2\columnwidth]{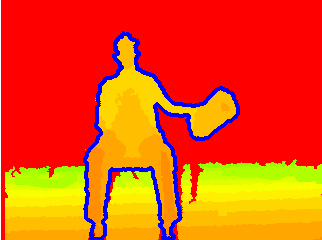}}%
\hspace{0.01cm}
\subfigure[]{\label{fig:hatb}\includegraphics[width=0.2\columnwidth]{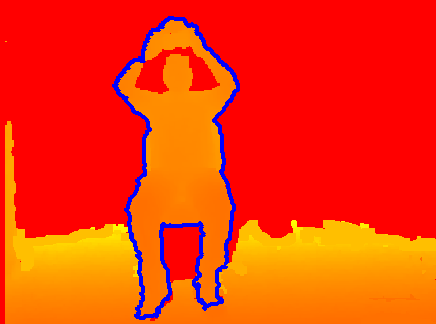}}
\hspace{0.01cm}
\subfigure[]{\includegraphics[width=0.2\columnwidth]{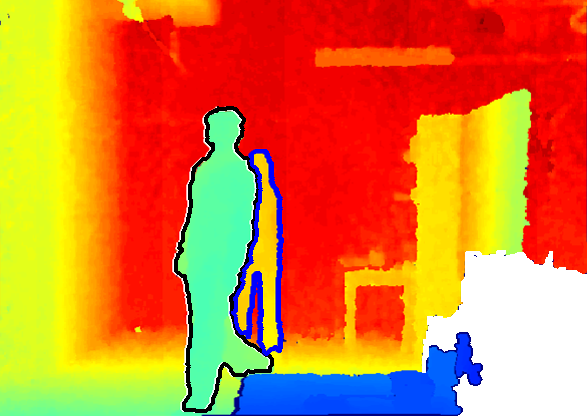}}
\caption{Object Detection Results during different activities: (a) Tennis serve, (b)Person Lifting Box, (c) Person Wearing Hat, and (d) Person Walking in a room}
\label{fig:detectionResultsSetA}
\end{figure}

We tested the proposed method on all datasets. Figure \ref{fig:detectionResultsSetA} shows the snapshots of the object detection while Table \ref{table:perfDataset} summarizes the accuracy of the proposed method on different datasets. From Table \ref{table:perfDataset}, we can see that the average accuracy of the proposed method is $\approx 93\%$. This clearly indicates that the proposed method is able to detect objects (humans, boxes, etc.) in most of the cases. The performance of the proposed method is low in case of IROS \cite{Spinello2011:eke}, Princeton \cite{Shuran2013:eke} and RWTH\cite{Rafi2015:eke} dataset. Since the proposed method detect and track an object by analysing its motion, it is not able to detect the objects which are stationary resulting in low accuracy.

\begin{table}[b!]
\small
\centering
\caption{Object detection results on different datasets}
\begin{tabular}{|c|c|c|c|}
\hline
Dataset & Precision & Recall & $F_1$ score \\
\hline
MSRAction3D & 100\% & 100\% & 100\% \\
\hline
MSRDailyActivity3D & 100\% & 100\% & 100\% \\
\hline
UT Austin & 100\% & 100\% & 100\% \\
\hline
IROS & 91\% & 84\% & 87.36\% \\
\hline
Princeton & 96\% & 81\% & 87.86\% \\
\hline
RWTH & 93\% & 84\% & 88.27\% \\
\hline
\multicolumn{3}{|r|}{Avg. Accuracy} & 93.91\% \\
\hline
\end{tabular}
\label{table:perfDataset}
\end{table}

Further, we compare the performance of the proposed object detection method with Xia et al.'s \cite{Xia2013:eke} noise suppression method in Table \ref{table:detectionDifferent}. From Table \ref{table:detectionDifferent}, we can see that the proposed noise suppression method results in higher accuracy than Xia et al.'s \cite{Xia2013:eke} method.

\begin{table}[t!]
\small
\centering
\caption{Object detection results with different noise Suppression methods (average values across all datasets)}
\begin{tabular}{|c|c|c|c|}
\hline
\textbf{Noise Suppression Method} & None & Xia \cite{Xia2013:eke} & Ours \\
\hline
$F_1$ score & 77\% & 84\% & 93\% \\
\hline
\end{tabular}
\label{table:detectionDifferent}
\end{table}

\textbf{Comparison with related work:} Table \ref{table:perfCompareAustin} contains the quantitative comparison between the proposed method and the related work. From Table \ref{table:perfCompareAustin}, we can see that the proposed method outperform the methods proposed by Ikemura et al. \cite{Ikemura2011:eke} and Xia et al. \cite{Xia2011:eke} while is comparable with the performance of Spinello et al.'s \cite{Spinello2011:eke} method. Ikemura et al.'s \cite{Ikemura2011:eke} method is window-based and is able to detect people which are well centered in frame resulting in high false negatives and low accuracy. Xia et al.'s \cite{Xia2011:eke} method detect the person by identifying the head contour. In case of occlusions, the method is able to detect the head contour of the foreground object and hence, resulting in false negatives and low accuracy. Further, we compared the accuracy of the proposed method with Combo-HOD detector proposed by Spinello et al. \cite{Spinello2011:eke}. The accuracy of the proposed method is around 93\% while the accuracy of Combo-HOD detector proposed by Spinello et al. \cite{Spinello2011:eke} is around 97\%. Since the proposed method detect objects based on the movement, it is not able to detect the humans which are stationary, resulting in slightly less accuracy than Spinello et al.'s \cite{Spinello2011:eke} method.

\begin{table}[t!]
\small
\centering
\caption{Comparison with different object detection methods (average values across all datasets)}
\begin{tabular}{|c|c|c|c|c|}
\hline
Method & Ikemura \cite{Ikemura2011:eke} & Xia \cite{Xia2011:eke} & Spinello \cite{Spinello2011:eke} & Ours\\
\hline
$F_1$ score & 61\% & 73\% & 97\% & 93\% \\
\hline 
\end{tabular}
\label{table:perfCompareAustin}
\end{table}
\subsection{Object Tracking Results}
\begin{figure}[h]
\centering
\subfigure[]{\includegraphics[width=0.3\columnwidth]{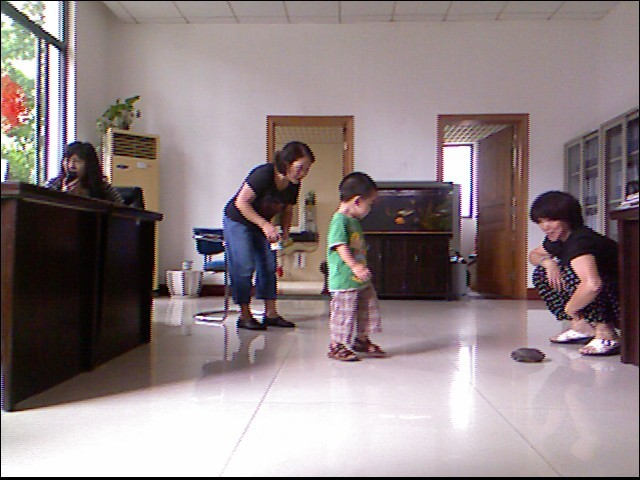}}
\subfigure[]{\includegraphics[width=0.3\columnwidth]{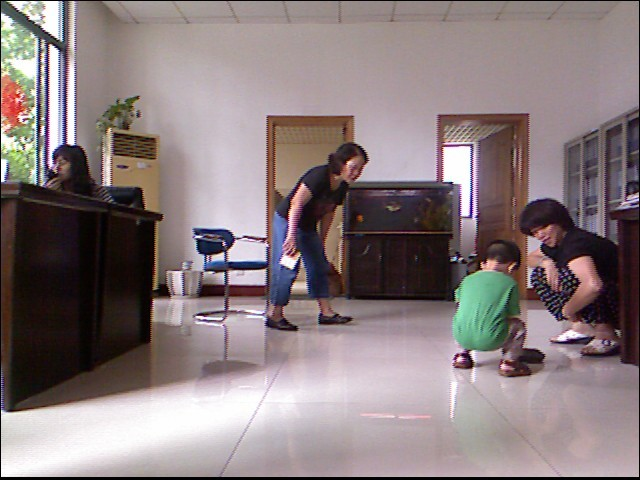}}
\subfigure[]{\includegraphics[width=0.3\columnwidth]{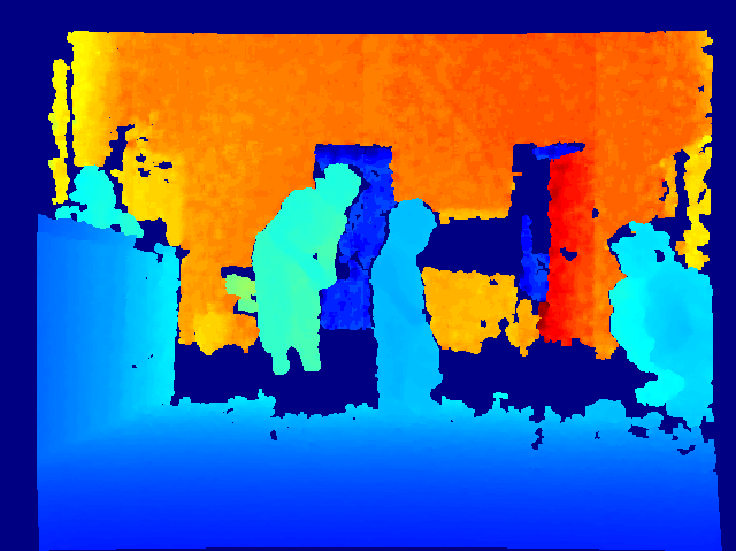}}
\hspace{0.05\columnwidth}
\subfigure[]{\includegraphics[width=0.3\columnwidth]{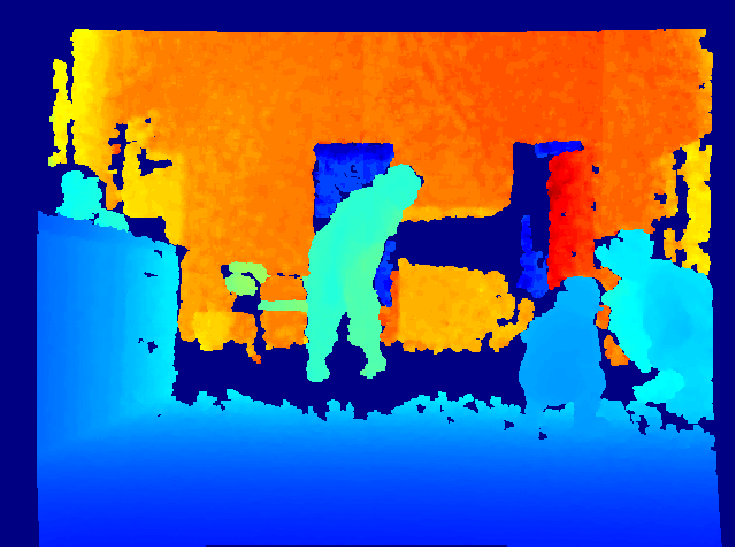}}
\subfigure[]{\includegraphics[width=0.3\columnwidth]{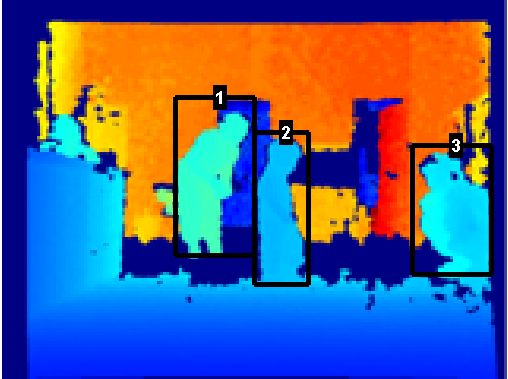}}
\subfigure[]{\includegraphics[width=0.3\columnwidth]{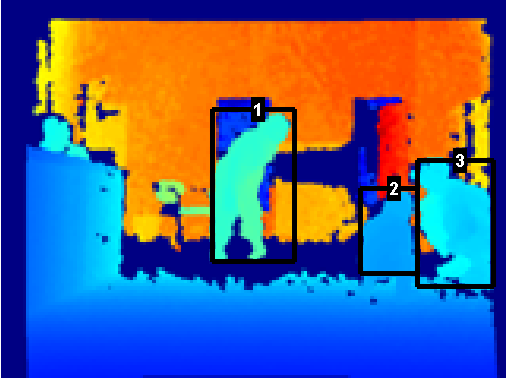}}
\caption{Tracking result on Princeton Dataset: (a-c) RGB frames, (d-f) depth frames corresponding to RGB frames (a-c), and (g-i) tracking results}
\label{fig:princetonExample}
\end{figure}

For object tracking, we tested the proposed method on all datasets. Figure \ref{fig:princetonExample} shows a snapshot of object tracking on one of the RGB-D videos in Princeton dataset. Video is captured inside a room where 4 people are present ( 3 adults and 1 kid). As we can see in Figure \ref{fig:princetonExample}, the proposed method is able to track only 3 people out of the 4 people inside the room. One of the adults present in the scene is stationary. The proposed method tracks the object based on motion of the object and hence, failed to detect one of the adults present inside the scene.

\begin{figure}[h]
\centering
\subfigure[]{\includegraphics[width=0.3\columnwidth]{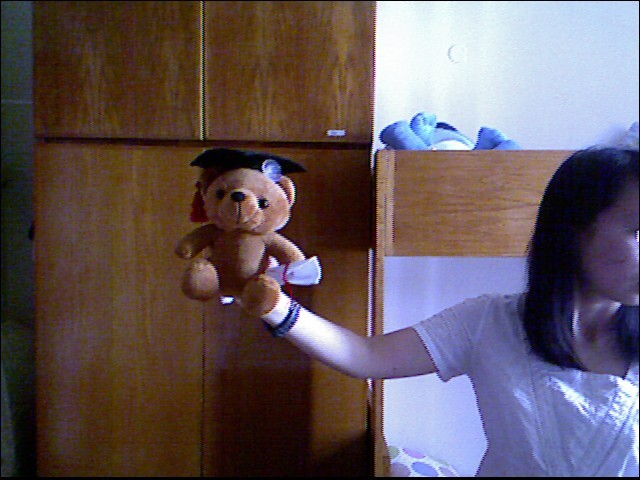}}
\subfigure[]{\includegraphics[width=0.3\columnwidth]{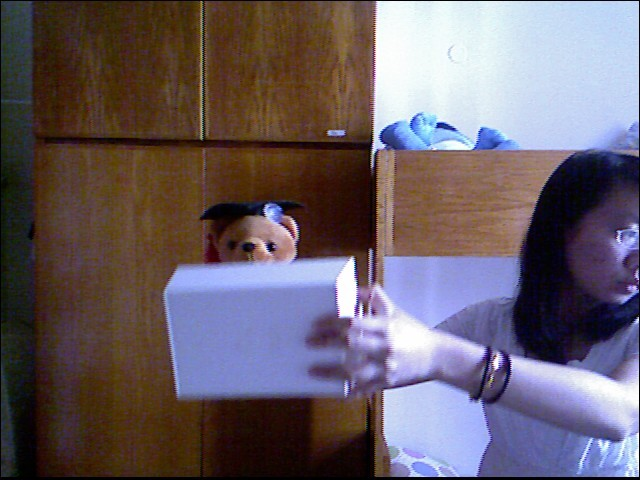}}
\subfigure[]{\includegraphics[width=0.3\columnwidth]{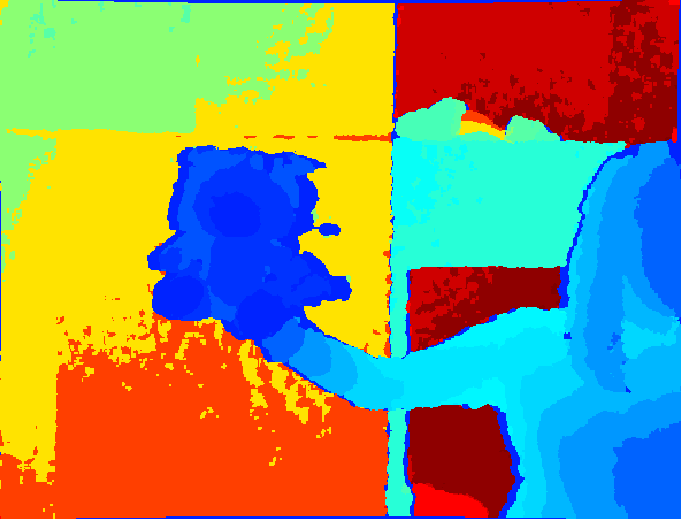}}
\hspace{0.05\columnwidth}
\subfigure[]{\includegraphics[width=0.3\columnwidth]{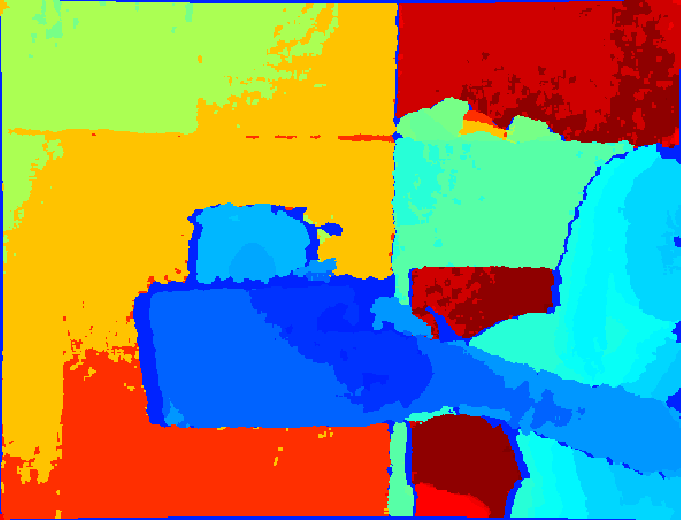}}
\subfigure[]{\includegraphics[width=0.3\columnwidth]{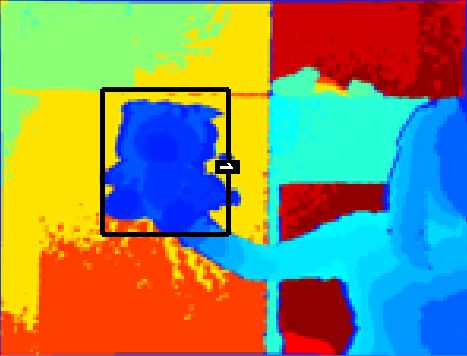}}
\subfigure[]{\includegraphics[width=0.3\columnwidth]{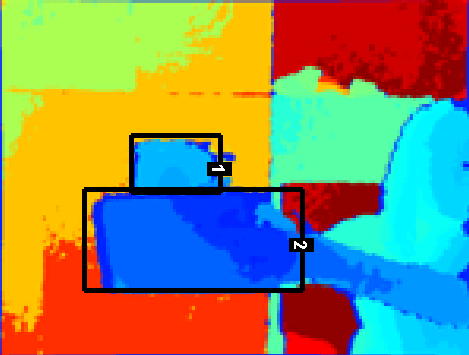}}
\caption{Results of tracking a bear: (a-b) RGB frames, (c-d) depth frames corresponding to RGB frames (a-b), and (e-f) tracking results}
\label{fig:princetonExampleTwo}
\end{figure}

Figure \ref{fig:princetonExampleTwo} shows a snapshot of tracking a bear. As we can see in Figure \ref{fig:princetonExampleTwo}, the proposed method is able to detect and track the bear as well as the box (with and without occlusion). However, the proposed method is not able to track the person. This is because the proposed method detect and track the objects which correspond to closed region $R$ and not the background region $R_B$ (see Section \ref{sec:proposedMethod} for details). The person in this video belongs to the background region $R_B$ and hence, the proposed method fails to detect and track the person. 

\begin{figure}[t!]
\centering
\includegraphics[width=0.7\columnwidth]{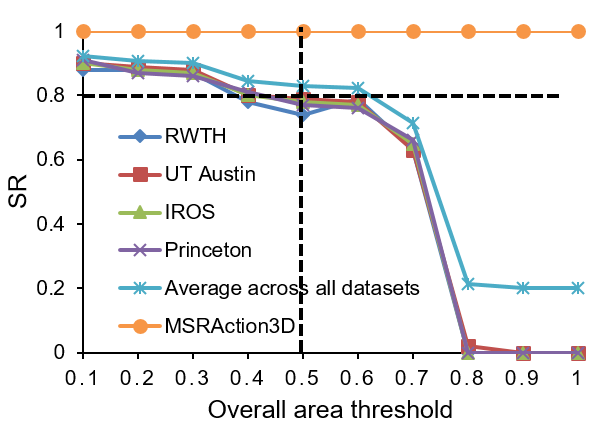}
\caption{Success rate ($SR$) vs. overlap area ($r$)}
\label{fig:successRate}
\end{figure}

For quantitative experiments, we computed overlapping ratio $r$ between tracked object and ground truth data (discussed in Section \ref{ssec:metrics}). Since RGB-D datasets are collected using different camera settings and under different conditions (such as indoor, outdoor, etc.), $r$ should be considered as a variable to conduct a fair study. We conducted experiments to adjust the value of $r$ and results are shown in Figure \ref{fig:successRate}. From Figure \ref{fig:successRate}, we can see that the average tracking SR of the proposed method is around 83\% across all datasets at $r = 0.5$.

Further, we compare the performance of the proposed object tracking method with Xia et al.'s \cite{Xia2013:eke} noise suppression method in Table \ref{table:trackingDifferent}. From Table \ref{table:trackingDifferent}, we can see that the proposed noise suppression method results in higher accuracy than Xia et al.'s \cite{Xia2013:eke} method.

\begin{table}[t!]
\small
\centering
\caption{Object tracking results with different noise Suppression methods}
\begin{tabular}{|c|c|c|c|}
\hline
\textbf{Noise Suppression Method} & None & Xia \cite{Xia2013:eke} & Ours \\
\hline
SR & 60\% & 74\% & 83\% \\
\hline
\end{tabular}
\label{table:trackingDifferent}
\end{table}

\textbf{Comparison with related work:} We compare the performance of the proposed tracking method against different methods in Table \ref{table:trackResults}. From Table \ref{table:trackResults}, we can see that the proposed method is more robust than the other methods. The higher accuracy of the proposed method against other methods is mainly because of the noise suppression. For instance, RGBOcc + OF \cite{Shuran2013:eke} method detects false features due to presence of noise, resulting in low accuracy. When we applied RGBOcc + OF \cite{Shuran2013:eke} method after suppressing the noise using the proposed method, we saw that the success rate increased from 75\% to 83\%.

\begin{table}[t!]
\small
\centering
\caption{Comparison with different object tracking methods (average values across all datasets)}
\begin{tabular}{|l|r|}
\hline
\textbf{Algorithm} & \textbf{SR} \\
\hline
RGBD HOG + Optical Flow (on depth data)\cite{Shuran2013:eke} & 75\% \\
\hline
RGB HOG + Optical Flow (on RGB data) \cite{Shuran2013:eke} & 52\% \\
\hline
Structured output tracking \cite{Hare2011:eke} & 46\% \\
\hline
Visual tracking \cite{Junseok2010:eke} & 42\% \\
\hline
Compressive tracking \cite{Kaihua2012:eke} & 40\% \\
\hline
Proposed method (on depth data) & \textbf{\textcolor{blue}{83}}\%\\
\hline
\end{tabular}
\label{table:trackResults}
\end{table}

\begin{figure}[h!]
\centering
\subfigure[]{\includegraphics[width=0.3\columnwidth, angle=90]{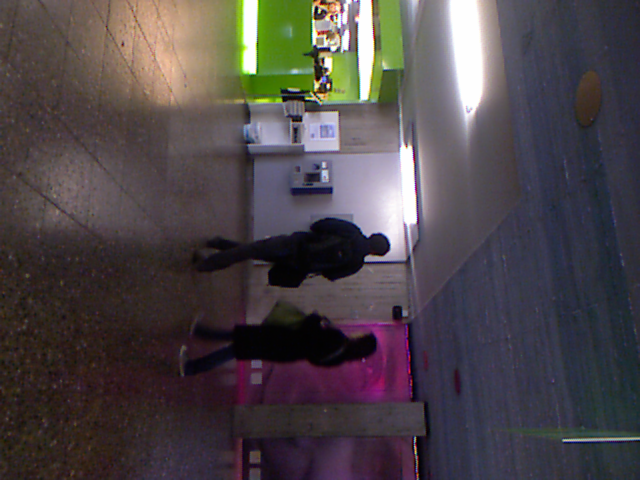}}
\hspace{1cm}
\subfigure[]{\includegraphics[width=0.3\columnwidth, angle=90]{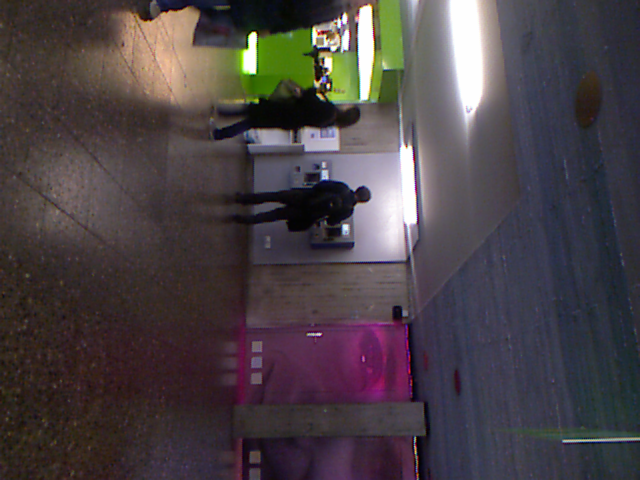}}
\hspace{1cm}
\subfigure[]{\includegraphics[width=0.3\columnwidth, angle=90]{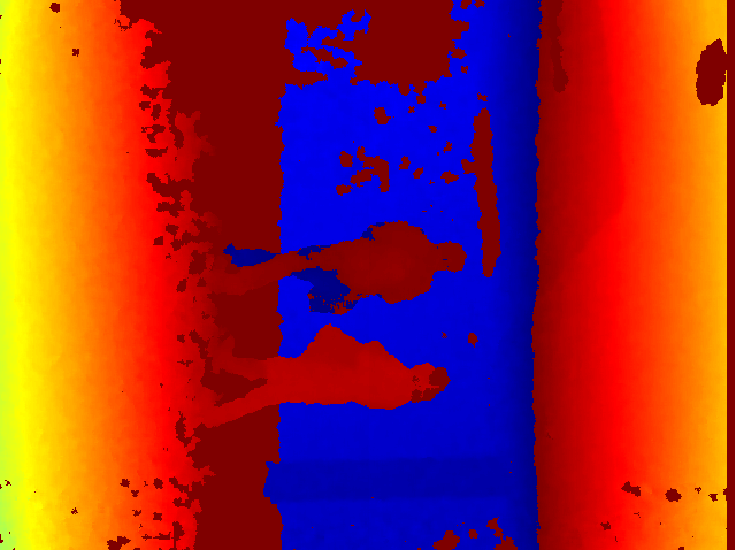}}
\subfigure[]{\includegraphics[width=0.3\columnwidth, angle=90]{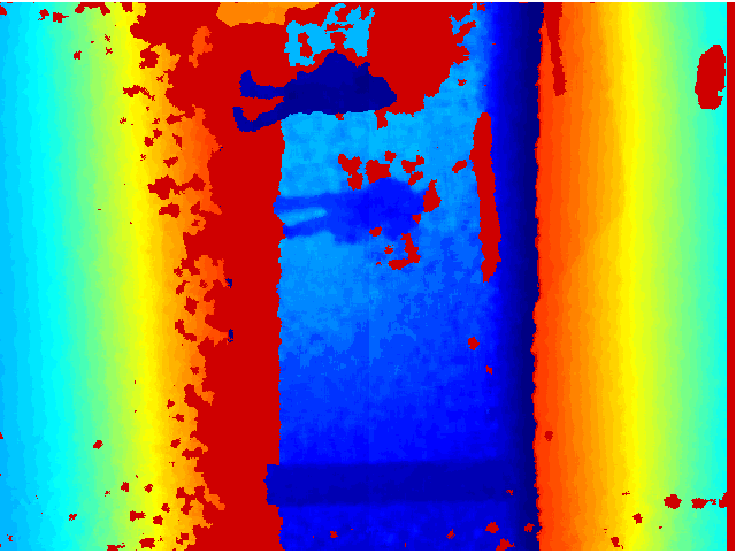}}
\hspace{1cm}
\subfigure[]{\includegraphics[width=0.3\columnwidth, angle=90]{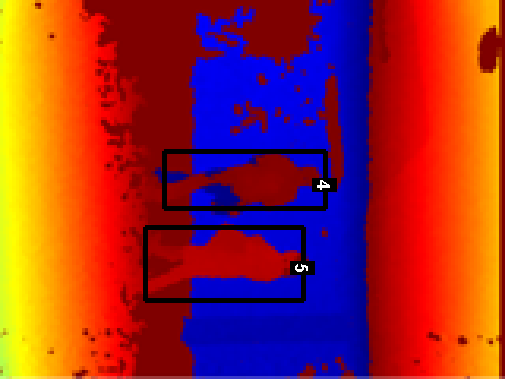}}
\hspace{1cm}
\subfigure[]{\label{fig:missingTrackex}\includegraphics[width=0.3\columnwidth, angle=90]{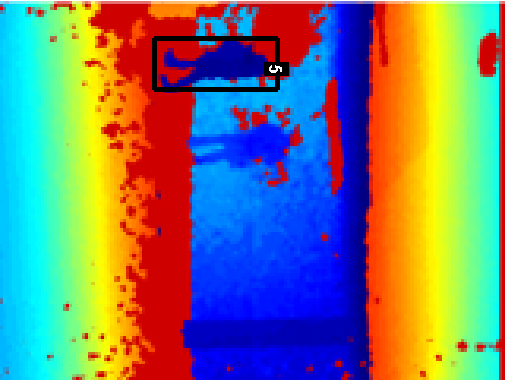}}
\caption{Example of lost track: (a-b) RGB frames, (c-d) depth frames, and (e-f) tracking results}
\label{fig:missingTrack}
\end{figure}

It is worth noting that the accuracy reported for object detection in Table \ref{table:perfDataset} and object tracking in Table \ref{table:trackResults} is different. This is because the proposed method is able to track only the motion part of the object. For instance, person might walk for sometime then does not move for sometime and then again start moving. In such a case, the proposed method tracks the person when he/she is moving. An example is shown in Figure \ref{fig:missingTrack} where two persons enter the scene (University Hall). One of them keeps walking while other goes to ATM machine and waits there for some time to complete his work. As we can see in Figure \ref{fig:missingTrack}, the proposed method is able to track both the person till they are moving. The proposed method subsequently lose the track of one of them as there is no movement (see Figure \ref{fig:missingTrackex}). Though we are able to detect both the persons in this scene, we are able to track these persons till they are walking. Such scenarios resulted in low SR.
\subsection{Impact on execution time}
\label{ssec:optimization}
The proposed method uses direction of object movement and weighted graph for optimizing the search area required for ROI detection and tracking. In case of optimization, we use direction of object movement and weighted graph for ROI detection and tracking while in case of without optimization, we search the entire frame to detect and track the ROI. Average execution time required per frame for noise reduction, ROI detection, and ROI tracking on different datasets by the proposed method without optimization and with optimization of search area are 888.6 and 540 milliseconds respectively i.e. the proposed optimized method is $1.65\times$ faster than the unoptimized method. Though the proposed optimization method significantly reduces the execution time, we have not seen any difference in the accuracy of the proposed method with and without optimization.
\section{Conclusion}
\label{sec:conclusion}
In this paper, we proposed a region graph based method for noise suppression, object detection, and object tracking using depth cameras. The experimental results show that the proposed method is able to detect and track the objects with and without occlusions. The proposed approach can be applied in different applications such as human activity recognition. 

The advantage of the proposed method can be summarized as: first, the proposed method does not require any training for ROI detection and tracking. Second, the proposed method uses direction of object movement and weighted graph for ROI detection and tracking. This helps in optimizing the searching area for detection and tracking. The limitations of the proposed method are: (i) it is not able to do re-identification of objects and (ii) object detection and tracking is not in real-time. RGB-based tracking systems seem to be more robust towards object re-identification, though their overall accuracy for tracking is lower. In future, we will try to address these limitations by: (i) complementing the proposed method with RGB data for person re-identification and (ii) utilizing GPU's for speeding-up performance of the proposed method.

\section{Acknowledgements}
This material is based upon work supported partially by the National Science Foundation under Grant No. 1012975. Any opinions, findings, and conclusions or recommendations expressed in this material are those of the author(s) and do not necessarily reflect the views of the National Science Foundation.

\bibliographystyle{abbrv}
{\small
\bibliography{wacvPaper}}
\end{document}